\documentclass{ifacconf}

\usepackage{tabularx}
\usepackage{multirow}
\usepackage{longtable,tabularx}
\usepackage{graphicx}     
\usepackage{wrapfig}
\usepackage{textcomp}
\usepackage{xcolor}
\usepackage{parskip}
\usepackage{amsmath,amssymb,amsfonts,bm}  
\usepackage{mathrsfs}
\usepackage{siunitx}
\usepackage{enumitem}
\usepackage{natbib}
\usepackage{algorithm}
\usepackage{algorithmicx}
\usepackage{algpseudocode}
\usepackage{nomencl}
\newtheorem{theorem}{Theorem}
\begin{document}
\begin{frontmatter}

\title{Model-Free and Real-Time Unicycle-Based Source Seeking with Differential Wheeled Robotic Experiments} 

\author[1]{Ahmed A. Elgohary}
\author[2]{Sameh A. Eisa}
\author[3]{Shivam Bajpai}

\address[1]{PhD Student, Department of Aerospace Engineering and Engineering Mechanics, University of Cincinnati, OH, USA (e-mail: elgohaam@mail.uc.edu)}

\address[2]{Assistant Professor, Department of Aerospace Engineering and Engineering Mechanics, University of Cincinnati, OH, USA (e-mail: eisash@ucmail.uc.edu)}

\address[3]{MS Student, Department of Aerospace Engineering and Engineering Mechanics, University of Cincinnati, OH, USA (e-mail: bajpaism@mail.uc.edu)}

\begin{abstract}                Many autonomous robots aimed at source-seeking are studied, and their controls designed, using unicycle modeling and formulation. This is true not only for model-based controllers, but also for model-free, real-time control methods such as extremum seeking control (ESC). In this paper, we propose a unicycle-based ESC design applicable to differential wheeled robots that: (1) is very simple design, based on one simple control-affine law, and without state integrators; (2) attenuates oscillations known to persist in ESC designs (i.e., fully stop at the source); and (3) operates in a model-free, real-time setting, tolerating environmental/sensor noise. We provide simulation and real-world robotic experimental results for fixed and moving light source seeking by a  differential wheeled robot using our proposed design. Results indicate clear advantages of our proposed design when compared to the literature, including attenuation of undesired oscillations, improved convergence speed, and better handling of noise.         
\end{abstract}


\end{frontmatter}

\section{Introduction}

Extremum Seeking Control (ESC) is a model-free, real-time adaptive control strategy designed to steer dynamical systems toward the extremum (maximum or minimum) of an objective function, even when the explicit form of that function is unknown (\cite{ariyur2003real,scheinker2024100}). ESC schemes are particularly attractive across various fields due to their minimal information requirements: they rely solely on applying perturbations to system parameters or inputs, and measuring the corresponding outputs of the objective function. Using these measurements in a feedback loop, ESC algorithms iteratively adjust the system input or parameter values to derive it towards the extremum, as can be seen in many applications -- see for example \cite{elgohary2025extremum,enatural_hovering2024,moidel2024reintroducing,pokhrel2024extremum}.

The problem of source-seeking has been widely studied due to its importance in applications such as environmental monitoring, chemical plume tracking, search-and-rescue missions, and robotic optimization tasks (e.g., \cite{chemical_application,rescue}). It typically involves steering an autonomous system (e.g., a mobile robot) towards a source that maximizes or minimizes the intensity of a physical field (e.g., heat, light, or chemical concentration), using only on-board sensor measurements. Traditional methods such as PID and model predictive control (MPC) (e.g., \cite{source_seeking1,source_seeking2}) require accurate models of both the robot and environment, which may be difficult to obtain in real-world conditions. In contrast, recent works have focused on model-free, real-time ESC strategies, which eliminate the need for explicit modeling and offer robust performance in uncertain environments for source-seeking, including successful experimental demonstrations -- see for example \cite{grushkovskaya2018family,ECC_2024,yilmaz2024unbiased}.

\textbf{Motivation.} ESC provides a powerful tool for real-time, model-free source-seeking, relying only on local measurements of a scalar signal and without requiring knowledge of the system dynamics or field model. However, practical implementations of ESC often suffer from persistent oscillations in control inputs (e.g., linear and angular velocities), which can degrade convergence accuracy and prevent the system from stopping precisely at the source (\cite{ghods2011extremum}). To resolve said persistent oscillations problem, some recent approaches have emerged (e.g., \cite{pokhrel2023control,grushkovskaya2024step,yilmaz2023exponential,yilmaz2024unbiased}). In our recent work \cite{ECC_2024}, we experimentally demonstrated the ability to achieve model-free source seeking with attenuated oscillation; hence, validating our approach in \cite{pokhrel2023control}. Let us consider the simple control law
\begin{equation}\label{eq:simple_law}
    \dot{x} = u = f(x) u_1 + u_2,
\end{equation}
where $f(x)$ is an unknown scalar field and $u_1, u_2$ are excitation signals. The law \eqref{eq:simple_law} approximates gradient flow without requiring knowledge of $f(x)$ or its gradient (\cite{durr2013lie}). Despite its theoretical appeal, real-world applications of law \eqref{eq:simple_law} have been limited. For instance, previous theoretical and experimental studies (\cite{grushkovskaya2018class,grushkovskaya2018family}) argued for its limitations and poor performance, due to for example, non-attenuating oscillations.

\textbf{Contribution.} In our recent study (\cite{ECC_2024}), we demonstrated for the first time, including experimentally, that the simple ESC law \eqref{eq:simple_law} can be used as a basis for single-integrator-based ESC design for model-free, real-time source-seeking with attenuating oscillations. In this paper, we extend those results to the more practical and widely used unicycle formulation, particularly for differential-wheeled robots. Specifically, we propose and validate a novel unicycle-based ESC design that: (i) does not require state integrators; (ii) enables oscillation attenuation as the robot approaches the source; (iii) improves convergence speed; and (iv) incorporates the Geometric Extended Kalman Filter (GEKF) (\cite{pokhrel2023gradient}) to enhance performance in noisy environments. The proposed method is simple, computationally efficient, and experimentally validated using the TB3 robot platform. Moreover, it significantly outperformed our recent single-integrator robotic results in \cite{ECC_2024}.

\section{Background}\label{section:II}
In this section, we briefly outline the essential background and theoretical preliminaries from our earlier work~(\cite{pokhrel2023control}), which form the base for the main results presented in the next section. Let us start with the class of control-affine ESC systems (\cite{durr2013lie,pokhrel2023higher}):
\begin{equation}\label{eqn:ESC_CA}
    \dot{\bm{x}} = \bm{b_d}(t, \bm{x}) + \sum\limits_{i=1}^{m} \bm{b_i}(t, \bm{x}) \sqrt{\omega} u_i( \omega t),
\end{equation}
where $\bm{x} \in \mathbb{R}^n$ is the state space vector, $\bm{b}_d$ the drift vector field, $m$ is the number of control inputs, $\bm{b}_i$ the control vector fields, $u_i$ are the control inputs, and $\omega \in (0, \infty)$ is a frequency parameter. Control-affine ESC systems in the form of \eqref{eqn:ESC_CA} can be approximated by the Lie Bracket System (LBS) approximation (\cite{durr2013lie}). In fact, very recently, we have been able to prove that LBS approximations to control-affine systems (including ESC) are, in fact, higher-order averaging themselves, i.e. they follow the traditional averaging analysis (\cite{pokhrel2023higher}). Now, the control-affine ESC system in \eqref{eqn:ESC_CA} can be approximated and represented by the LBS (\cite{durr2013lie,pokhrel2023higher}):
\begin{equation}\label{eqn:Lie_back_unicycle}
    \dot{\bm{z}} = \bm{b_d}(t, \bm{z}) + \sum_{\substack{i=1\\j=i+1}}^m [\bm{b_i}, \bm{b_j}](t, \bm{z}) \nu_{j,i}(t),
\end{equation}
with $\nu_{j,i}(t) = \frac{1}{T} \int_0^T u_j(t, \theta) \int_0^\theta u_i(t, \tau) d\tau d\theta$, and the Lie bracket $[\bm{b_i}, \bm{b_j}] := \frac{\partial \bm{b_j}}{\partial \bm{x}} \bm{b_i} - \frac{\partial \bm{b_i}}{\partial \bm{x}} \bm{b_j}$. where $\bm{b_i}, \bm{b_j}: \mathbb{R} \times \mathbb{R}^n \to \mathbb{R}^n$ are continuously differentiable vector fields. We now focus on a subclass of control-affine ESCs, which describe the ESC law \eqref{eq:simple_law} and many others ESC laws in literature (\cite{grushkovskaya2018class}):
\begin{equation}\label{eqn:ESC}
    \dot{x} = b_1(f(x)) u_1 + b_2(f(x)) u_2,
\end{equation}
where $f(x)$ is the unknown objective function, and the excitation inputs are $u_1 = a\sqrt{\omega} \hat{u}_1(\omega t)$ and $u_2 = a\sqrt{\omega} \hat{u}_2(\omega t)$.

The corresponding LBS to an ESC of the form \eqref{eqn:ESC} is given by (\cite{grushkovskaya2018class}):
\begin{align}\label{eqn:Lie}
    \dot{z} &= -\nu_{2,1} \nabla f(z) b_0(f(z)), \\
    \text{where} \quad b_0(z) &= b_2(z) \frac{db_1(z)}{dz} - b_1(z) \frac{db_2(z)}{dz}. \nonumber
\end{align}
In higher dimensions, \eqref{eqn:ESC} generalizes to:
\begin{equation}\label{eqn:ESC_multi}
    \dot{\bm{x}} = \sum_{i=1}^n \left( b_{1i}(f(\bm{x}))u_{1i} + b_{2i}(f(\bm{x}))u_{2i} \right) e_i,
\end{equation}
with the LBS:
\begin{equation}\label{eqn:Lie_multi}
    \dot{\bm{z}} = -\sum_{i=1}^n \nu_{2i,1i} \frac{\partial f(\bm{z})}{\partial z_i} b_{0i}(f(\bm{z})) e_i.
\end{equation}
Now, assuming one has access to the estimated LBS in (\ref{eqn:Lie_multi}) using measurements of the objective function \( f(\bm{x}) \), which depends on the gradient of the objective function, then the estimated LBS, denoted as \( \hat{\bm{z}} \), can be expressed as \( \dot{\hat{\bm{z}}} = -\sum_{i=1}^n \left( \nu_{2i,1i} \frac{\partial f(\bm{z})}{\partial z_i} b_{0i}(f(\bm{z})) + \eta_i(t) \right) e_i = \bm{J}(t, \bm{z}) \), where \( \eta_i(t) \) represents the error term accounting for measurement noise. A class of control-affine ESC systems has been proposed in (\cite{pokhrel2023control}) that couples the ESC system in (\ref{eqn:ESC_multi}) with an adaptation law for the amplitude of the escitation signal that relies on the estimated LBS. Said ESC class is expressed as follows:
\begin{align}
    \dot{\bm{x}} &= \sum_{i=1}^n \left( b_{1i}(f(\bm{x})) \sqrt{\omega} a_i(t) \hat{u}_{1i} + b_{2i}(f(\bm{x})) \sqrt{\omega} a_i(t) \hat{u}_{2i} \right) e_i, \label{eqn:generalizedSystem} \\
    \dot{\bm{a}} &= \sum_{i=1}^n \left( -\lambda_i \left( a_i(t) - J_i(t, \bm{x}) \right) \right) e_i, \label{eqn:law}
\end{align}
where  \( J_i(t, \bm{x}) \) represents the right-hand side of the estimated LBS, \( a_i \in \mathbb{R} \) represents the amplitude of the input signal, \( \lambda_i > 0 \in \mathbb{R} \) are tuning parameters, and \( u_i = a_i \hat{u}_i \). The ESC class in \eqref{eqn:generalizedSystem}-\eqref{eqn:law} provides designs guaranteed to attenuate in oscillations as shown in (\cite{pokhrel2023control}). The applicability/functionality of the ESC class in \eqref{eqn:generalizedSystem}-\eqref{eqn:law} relies on the estimations of the gradient of the unknown objective function \( f(\bm{x}) \) (i.e., $\nabla f(\bm{x})$), which in turn provides \( J_i(t, \bm{x}) \). Said estimations is guranteed to have high precision with an error that vanishes as \( t \to \infty \) as in (\cite{pokhrel2023control}). We assume the following assumptions from (\cite{pokhrel2023control}):
\noindent\\
\textbf{A1} $b_{ji}, b_{ji}\in C^2: \mathbb{R} \to \mathbb{R}$, and for a compact set $\mathscr{C} \subset \mathbb{R}$, there exist $A_1, ..., A_3 \in [0,\infty)$ such that $|b_{j}(x)|\leq A_1,
\left| \frac{\partial b_j(x)}{\partial x} \right|\leq A_2,
\left| \frac{\partial [{b_j},{b_k}](x)}{\partial x} \right| \leq A_3$ for all $x\in \mathscr{C},\ i={1,2};\; j={1,2};\; k={1,2}$. \\
\textbf{A2} $\hat{u}_{1i}, \hat{u}_{2i}: \mathbb{R} \times \mathbb{R} \to \mathbb{R} , i=1,2$, are measurable functions. Moreover, there exist constants $M_i \in (0,\infty)$ such that 
$\sup_{\omega t \in \mathbb{R}}|\hat{u}_i(\omega t)|\leq M_i$, and 
$\hat{u}_i(\omega t + T)=\hat{u}_i(\omega t)$, and 
$\int_0^T \hat{u}_i(\tau) d\tau = 0$, with $T \in (0,\infty)$ for all $\omega t \in \mathbb{R}$. \\[0.5em]
\textbf{A3} There exists an ${x}^* \in \mathscr{C}$ such that $\nabla f({x}^*)=0,\ \nabla f({x})\ne 0$ for all ${x}\in \mathscr{C}\backslash \{{x}^*\};\ f({x}^*)=f^* \in \mathbb{R}$ is an isolated extremum value.\\
\textbf{A4} Let ${\eta}_i(t)$ denote the estimation error of ${J}_i(t,\bm{x})$. Then ${\eta}_i(t): \mathbb{R} \rightarrow \mathbb{R},\ i=1,\ldots,n$, is a measurable function for which there exist constants $\theta_0, \epsilon_0 \in (0,\infty)$ such that $|{\eta}_i(t_2) - {\eta}_i(t_1)| \le \theta_0 |t_2 - t_1|$ for all $t_1, t_2 \in \mathbb{R}$, and $\sup_{t \in \mathbb{R}} |{\eta}_i(t)| \le \epsilon_0$. Moreover, $\lim_{t \to \infty} {\eta}_i(t) = 0$.
We recall the following theorem from~(\cite{pokhrel2023control}) which establishes the stability properties of the ESC system (\ref{eqn:generalizedSystem}),(\ref{eqn:law})
\begin{theorem}\label{thm:esc_lbs}
Let assumptions A1-A4 be satisfied and suppose that a compact set $\mathscr{C}$ is locally (uniformly) asymptotically stable for (\ref{eqn:Lie_multi}). Then $\mathscr{C}$ is locally practically (uniformly) asymptotically stable for (\ref{eqn:ESC_multi}).
\end{theorem}

\section{Main Results}\label{Sec:Main_results}
In this section, we provide our proposed unicycle-based ESC design, applicable to autonomous robotic systems, particularly differential wheeled robots. 

\subsection{Structure of the Proposed Unicycle-Based ESC Design}
\begin{figure}[ht]
\centering
\includegraphics[width=0.4\textwidth]{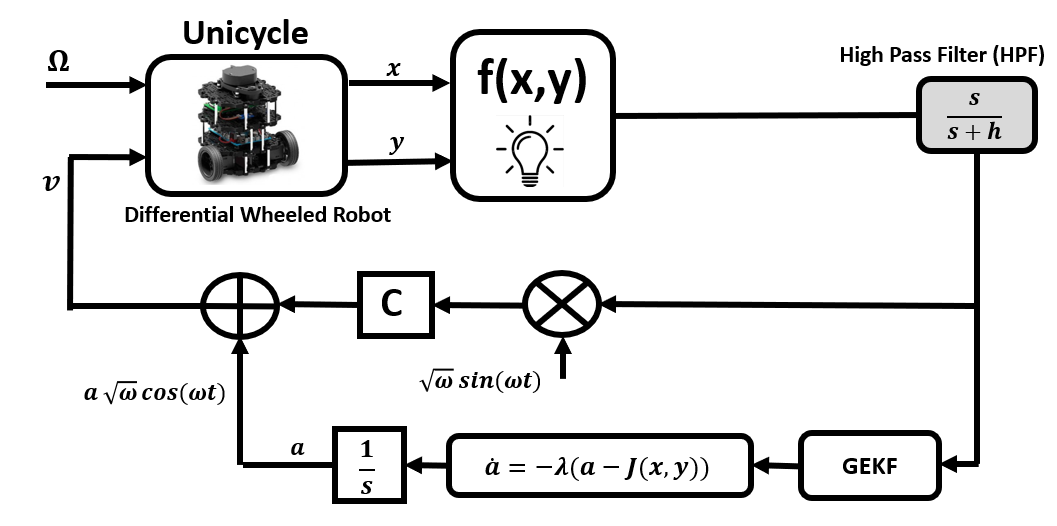}
\caption{The proposed unicycle-based ESC design}
\label{fig:ESC_scheme}
\end{figure}

Let us now construct the rationale of our proposed design in three steps. Firstly, we start by reconsidering 
the unicycle dynamics described in \cite{durr2013lie} but with only a single control input that is the linear velocity \( v \) (i.e., $u=v$) and a fixed angular velocity \( \Omega \). Secondly, we propose using the very simple ESC law in \eqref{eq:simple_law} to determine the update of the linear velocity \( v \). We also apply a high-pass filter (HPF) to the measurement of the objective function $f(x,y)$ to improve the transient behavior and the accuracy of sensor measurements. That is, $u=v=  ( f(x, y) - eh )u_1 + u_2$, where $u_1$ and $u_2$ are the ESC perturbation/excitation signals (usually are taken as sinusoidal signals, but other forms are allowed as long as they satisfy assumption A2), $e$ is the resulting state from using the HPF with constant $h$ that is greater than or equals to zero. Thirdly, we turn the unicycle dynamics with the aforementioned considerations to be effectively a member of our recent ESC class proposed in \cite{pokhrel2023control} to guarantee attenuating oscillations and better performance due to the use of GEKF that enables better handling of sensor noise and complete stopping at the source if it is a fixed one. That is, our unicycle-based ESC follows the state space representation in    \eqref{eqn:generalizedSystem}-\eqref{eqn:law}, incorporating an adaptation mechanism that reduces the amplitude $a$ of the perturbation/excitation signals $u_1$ and $u_2$. Hence, the linear velocity $v$ will automatically have lesser perturbations as it gets closer to the terminal/desired source the robot is seeking (i.e., $v$ perturbs as needed and vanishes at the source the robot is seeking). However, our proposed design is proven to be capable of tracking moving source as well; this will be shown in Section 4. The proposed unicycle-based ESC design operates in a planar mode (\(x\) and \(y\) coordinates) with $u_1=sin(\omega t)$ and $u_2=cos(\omega t)$, where $\omega$ is the perturbation/excitation frequency. The proposed design has the following state space representation on the form of systems \eqref{eqn:generalizedSystem},\eqref{eqn:law}:
\begin{equation}
\begin{aligned}
\dot{x} &= \left( \left( f(x, y) - eh \right) c \sqrt{\omega} \sin(\omega t) + a \sqrt{\omega} \cos(\omega t) \right) \cos \Omega t, \\
\dot{y} &= \left( \left( f(x, y) - eh \right) c \sqrt{\omega} \sin(\omega t) + a \sqrt{\omega} \cos(\omega t) \right) \sin \Omega t, \\
\dot{e} &= -he + f(x, y), \\
\dot{a} &= -\lambda \left( a - J(x, y) \right),
\end{aligned}
\label{eqn:extended_system}
\end{equation}
where $c$ and $\lambda$ are  tuning constants. Furthermore, the proposed unicycle-based ESC design is provided in Figure \ref{fig:ESC_scheme}.
Additionally, the LBS corresponding to the first three equations in (\ref{eqn:extended_system}) that will be used for the GEKF implementation (to determine $J(x,y)$), is based on (\ref{eqn:Lie_multi}) and is obtained as follows:
\begin{equation}
\begin{bmatrix}
\dot{\bar{x}} \\
\dot{\bar{y}} \\
\dot{\bar{e}}
\end{bmatrix}
=
\frac{1}{2}
\begin{bmatrix}
c a \nabla_x f(x, y) \cos^2(\Omega t) \\
c a \nabla_y f(x, y) \sin^2(\Omega t) \\
2(-\bar{e} h + f(x, y))
\end{bmatrix}.
\label{eq:LBS_ESC}
\end{equation}

For successful implementation of the proposed design (Figure \ref{fig:ESC_scheme}), we need to derive the GEKF for the proposed design. As shown in~\cite{pokhrel2023control}, the validity of Theorem~\ref{thm:esc_lbs} relies on accurate estimation of the gradient terms, specifically \( J(x,y) \), with the requirement that the estimation error vanishes as \( t \to \infty \). This condition was successfully satisfied using GEKF, a novel estimation framework introduced in~\cite{pokhrel2023gradient}. For full details on the GEKF design, its theoretical foundations, and derivations, we refer the reader to~\cite{pokhrel2023control,pokhrel2023gradient}. The proposed GEKF will have five states. The first two states of the proposed GEKF will be the first two enteries of the right-hand side of the LBS in (\ref{eq:LBS_ESC}). Then, the third and fourth states will be the derivatives of the first and second states themselves, respectively. The fifth state will be the third entry of the right-hand side of the LBS in (\ref{eq:LBS_ESC}). As a result, the proposed GEKF states are defined as follows:
\begin{align}
    \bar{\bm{X}}=\begin{bmatrix}
    \bar{x}_1\\
    \bar{x}_2\\
    \bar{x}_3\\
    \bar{x}_4\\
    \bar{x}_5
    \end{bmatrix}=\begin{bmatrix}
    \frac{c a}{2} \nabla_{x} f(x,y)\cos^2 (\Omega t)\vert _{t_1}\\
    \frac{ca}{2} \nabla_{y} f(x,y) \sin^2 (\Omega t)\vert _{t_1}\\
    \dot{\bar{x}}_1\\
     \dot{\bar{x}}_2\\
     (f(x, y) - eh)\vert _{t_1}\\
    \end{bmatrix}_{5 \times 1}.
    \label{eq:X_bar}
\end{align}
Using a constant velocity propagation model as in (\cite{ECC_2024}), we get:
\begin{align}\label{eqn:GEKFDynamics_eg3_row}
\dot{\bar{\bm{X}}}
  &=
  \begin{bmatrix}
      \bar{\bm{x}}_3 & \bar{\bm{x}}_4 & 0 & 0 & 0
  \end{bmatrix}^{\!\top}
  + \bm{\Omega}.
\end{align}
where \( \bm{\Omega} \) denotes the process noise vector modeled as a random variable, \( \bm{Q} \) represents the covariance matrix associated with the process noise (system noise), and the Jacobian matrix \( \bm{A} \), corresponding to the state dynamics, is computed as:
\begin{align}\label{eqn:matrixA_case3}
    \bm{A}=\begin{bmatrix}
    0 & 0 & 1 & 0 & 0\\
    0 & 0 & 0 & 1 & 0\\
    0 & 0 & 0 & 0 & 0\\
    0 & 0 & 0 & 0 & 0\\
    0 & 0 & 0 & 0 & 0\\
    \end{bmatrix}_{5 \times 5}.
\end{align}
For the measurement updates, the Chen-Fliess series expansion is utilized as detailed in \cite{pokhrel2023control,pokhrel2023gradient}. Using the Chen-Fliess series expansion and truncating the series after the first-order terms, we obtain:
{\small
\begin{equation}\label{eqn:measurementeqn3}
\begin{split}
    &(f(x, y) - eh)\big|_{t_2} = (f(x, y) - eh)\big|_{t_1} \\
    &\quad + \left(\nabla f(x, y) \cdot \mathbf{b}_1\right)\big|_{t_1} K \cos(\omega t) \\
    &\quad + \left(\nabla f(x, y) \cdot \mathbf{b}_2\right)\big|_{t_1} K \sin(\omega t) + \nu(t) \\
    &\text{where} \quad
    \mathbf{b}_1 = \begin{bmatrix}
        c(f(x, y) - eh)\cos(\Omega t) \\
        c(f(x, y) - eh)\sin(\Omega t) \\
        0
    \end{bmatrix},
    \mathbf{b}_2 = \begin{bmatrix}
        a\cos(\Omega t) \\
        a\sin(\Omega t) \\
        0
    \end{bmatrix},
\end{split}
\end{equation}}
and \( t_2 = t_1 + \Delta t \). Based on the approach in \cite{pokhrel2023control,pokhrel2023gradient}, \( K \) is defined as \( K = \frac{2}{\sqrt{\omega}} \sin\left(\frac{\omega \Delta t}{2}\right) \), where \( \Delta t \) denotes a small time step between measurements. It is critical to ensure that \( \Delta t \) is significantly smaller than the period of the input signal.
The residual terms are assumed to follow a Gaussian distribution representing measurement noise, denoted as \( \nu(t) \sim N(0, R) \), where \( R \) is the covariance matrix associated with the noise.
The measurement update equation can then be rewritten in terms of the GEKF states as:
\begin{equation*}
\begin{split}
    &(f(x, y) - eh)\big|_{t_2} = (f(x, y) - eh)\big|_{t_1} \\
    &\quad + \left( \nabla_x f(x, y) (f(x, y) - eh) c \cos(\Omega t) \right. \\
    &\quad + \left. \nabla_y f(x, y) (f(x, y) - eh) c \sin(\Omega t) \right) K \cos(\omega t) \\
    &\quad + \left( \nabla_x f(x, y) a \cos(\Omega t) 
    + \nabla_y f(x, y) a \sin(\Omega t) \right) K \sin(\omega t) \\
    &\quad + \nu(t) \\
    &= \bar{x}_5 
    + \frac{2K}{a \cos(\Omega t)} \bar{x}_1 \bar{x}_5 \cos(\omega t)
    + \frac{2K}{a \sin(\Omega t)} \bar{x}_2 \bar{x}_5 \cos(\omega t) \\
    &\quad + \frac{2K}{c \cos(\Omega t)} \bar{x}_1 \sin(\omega t)
    + \frac{2K}{c \sin(\Omega t)} \bar{x}_2 \sin(\omega t)
    + \nu(t).
\end{split}
\end{equation*}
where \( t = (t_1 + t_2) / 2 \). 
The Jacobian matrix associated with the measurement update equation is:
\begin{align*}
    \bm{C} = \begin{bmatrix}
        \frac{2K}{a \cos(\Omega t)} \bar{x}_5 \cos(\omega t) + \frac{2K}{c \cos(\Omega t)} \sin(\omega t) \\
        \frac{2K}{a \sin(\Omega t)} \bar{x}_5 \cos(\omega t) + \frac{2K}{c \sin(\Omega t)} \sin(\omega t) \\
        0 \\
        0 \\
        1 + \frac{2K}{a \cos(\Omega t)} \bar{x}_1 \cos(\omega t) + \frac{2K}{a \sin(\Omega t)} \bar{x}_2 \cos(\omega t)
    \end{bmatrix}^T.
\end{align*}
Now we are in a position to provide Algorithm \ref{alg:alg1}, which is an implementation algorithm for the proposed unicycle-based ESC design operating with GEKF as in Figure \ref{fig:ESC_scheme}. 
The GEKF operates by propagating the state estimate and covariance through the prediction step, followed by measurement updates whenever new sensor data is available. Since the operation of GEKF is necessary for the attenuation of oscillations (incurring adaptive perturbation/excitation) via reducing the amplitude $a$ (see Figure \ref{fig:ESC_scheme}), then we can effectively introduce a stopping condition when $a$ is smaller than a certain very small threshold $\epsilon$ (e.g., $\epsilon = 0.01$) as this indicates that the robot has practically reached the source. Said stopping condition will be applied via the only control input $v$ (the linear velocity). That is, if $a\leq \epsilon$, then the input is $v=0$, effectively halting its motion. Otherwise, the GEKF continues to estimate the system states. 
\begin{algorithm}
\caption{Unicycle-based Source Seeking Algorithm (Using GEKF)}
\label{alg:alg1}
{\small
\begin{algorithmic}
\State \textbf{Initialization:}
\State \textbf{•} Set the initial state estimate \( \bar{\bm{X}} \) in (\ref{eq:X_bar}).
\State \textbf{•} Select the output sample time \( \Delta t \), ensuring that it is lower than the sensor sampling rate.
\State \textbf{•} Select \( N \) as the number of discrete subintervals within \( \Delta t \) for improving the accuracy of the prediction step.
\State \textbf{•} Set the threshold value \( \varepsilon \) that determines when the algorithm stops (e.g., \( \varepsilon = 0.01 \)).

\State \textbf{Simulation Begins: (the system (\ref{eqn:extended_system}) is active)}
\State \textbf{•} Track the amplitude of the oscillatory input signal \( a \) (the last variable in (\ref{eqn:extended_system})) to assess whether the system dynamics need to continue in activation or stop.

\If{$a > \varepsilon$}
    \State \textbf{Start Estimating State \( \bar{\bm{X}} \):}
    \For{$i = 1$ to $N$} \text{(Prediction Step)}
        \State \textbf{•} \( \bar{\bm{X}} = \bar{\bm{X}} + \frac{\Delta t}{N} \dot{\bar{\bm{X}}} \)
        \State \textbf{•} \( \bm{P} = \bm{P} + \frac{\Delta t}{N} (\bm{A}\bm{P} + \bm{P}\bm{A}^T + \bm{Q}) \)
    \EndFor
    \State \textbf{Return:} \( \bar{\bm{X}} \) and \( \bm{P} \)
    \State \textbf{Measurement Update:}
    \State \textbf{•} When a new measurement is received:
    \State \textbf{•} \( \bm{C} = \frac{\partial \bm{h}}{\partial \bm{X}}(\bar{\bm{X}}) \)
    \State \textbf{•} \( \bm{L} = \bm{P} \bm{C}^T (\bm{R} + \bm{C} \bm{P} \bm{C}^T)^{-1} \)
    \State \textbf{•} \( \bm{P} = (\bm{I} - \bm{L} \bm{C}) \bm{P} \)
    \State \textbf{•} \( \bar{\bm{X}} = \bar{\bm{X}} + \bm{L}(\bm{y} - \bm{C} \bar{\bm{X}}) \)
\Else
    \State \textbf{Stop the System:}
    \State \textbf{•} Set the velocity \( v = 0 \)
\EndIf
\end{algorithmic}
}
\end{algorithm}
\begin{figure*}[t]
    \centering
    \begin{minipage}[t]{0.3\textwidth}
        \centering
        \includegraphics[width=\textwidth]{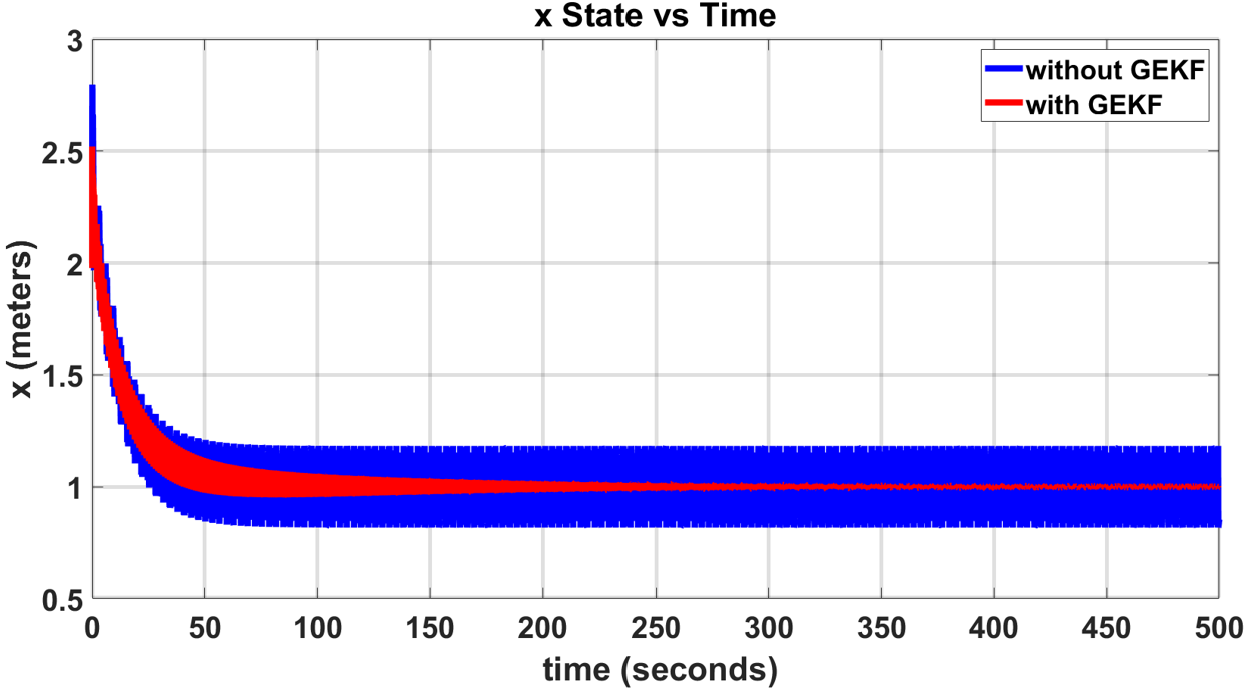}
    \end{minipage}
    \begin{minipage}[t]{0.3\textwidth}
        \centering
        \includegraphics[width=\textwidth]{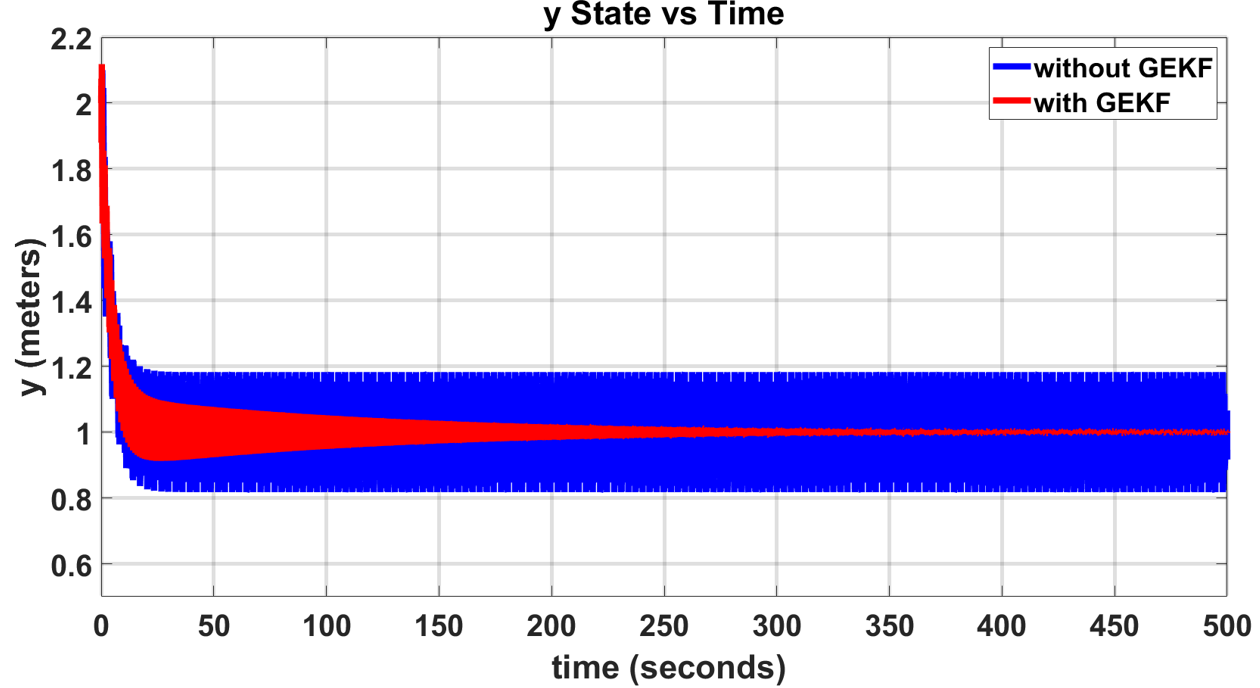}
        
    \end{minipage}
    \begin{minipage}[t]{0.3\textwidth}
        \centering
        \includegraphics[width=\textwidth]{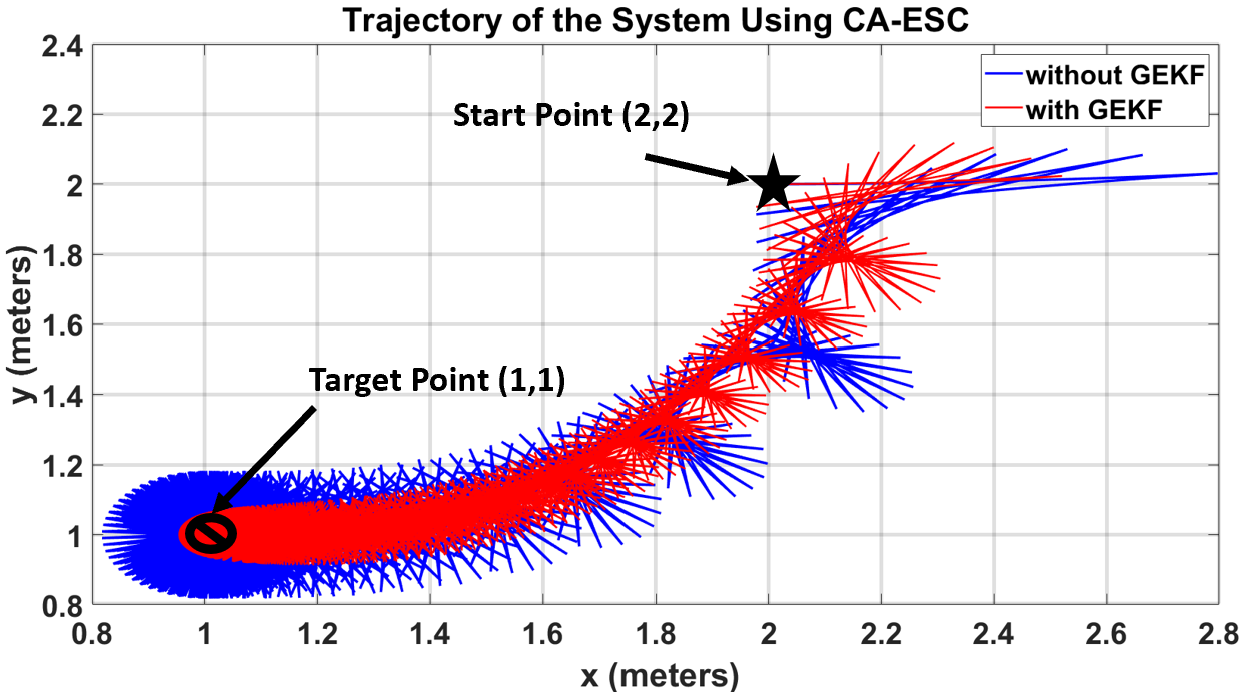}
        
    \end{minipage}
    \caption{Simulation results of \(x\), \(y\) states and trajectory. GEKF-based design (red) reduces oscillations vs. baseline (blue).}
    \label{fig:Simulation_R}
\end{figure*}
\begin{figure*}[t]
    \centering
    \begin{minipage}[t]{0.3\textwidth}
        \centering
        \includegraphics[width=\textwidth]{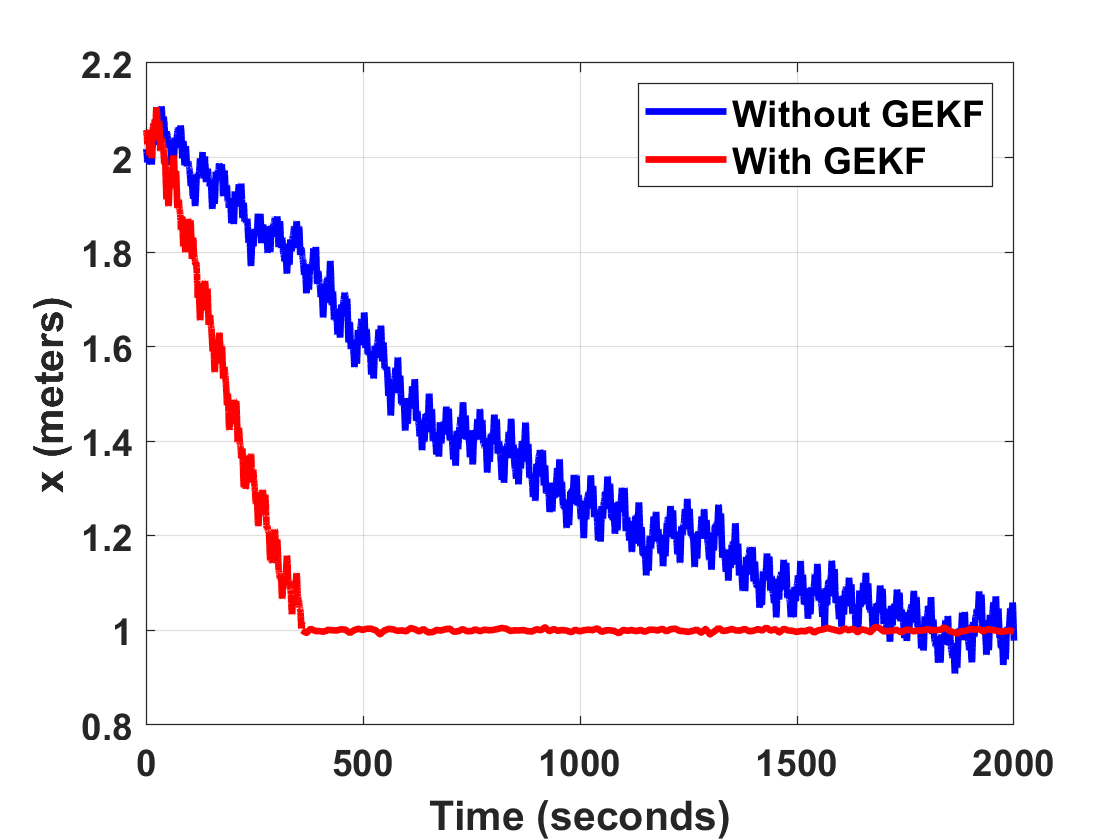}
        
    \end{minipage}
    \begin{minipage}[t]{0.3\textwidth}
        \centering
        \includegraphics[width=\textwidth]{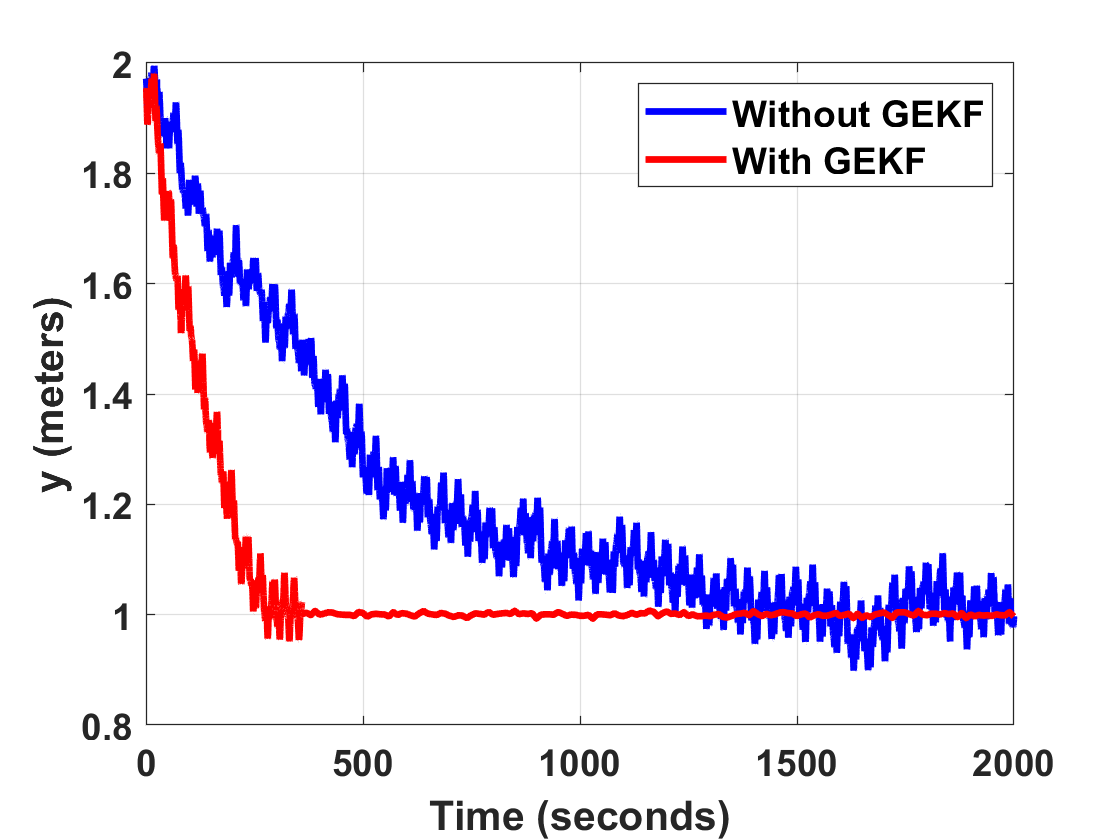}
        
    \end{minipage}
    \begin{minipage}[t]{0.3\textwidth}
        \centering
        \includegraphics[width=\textwidth]{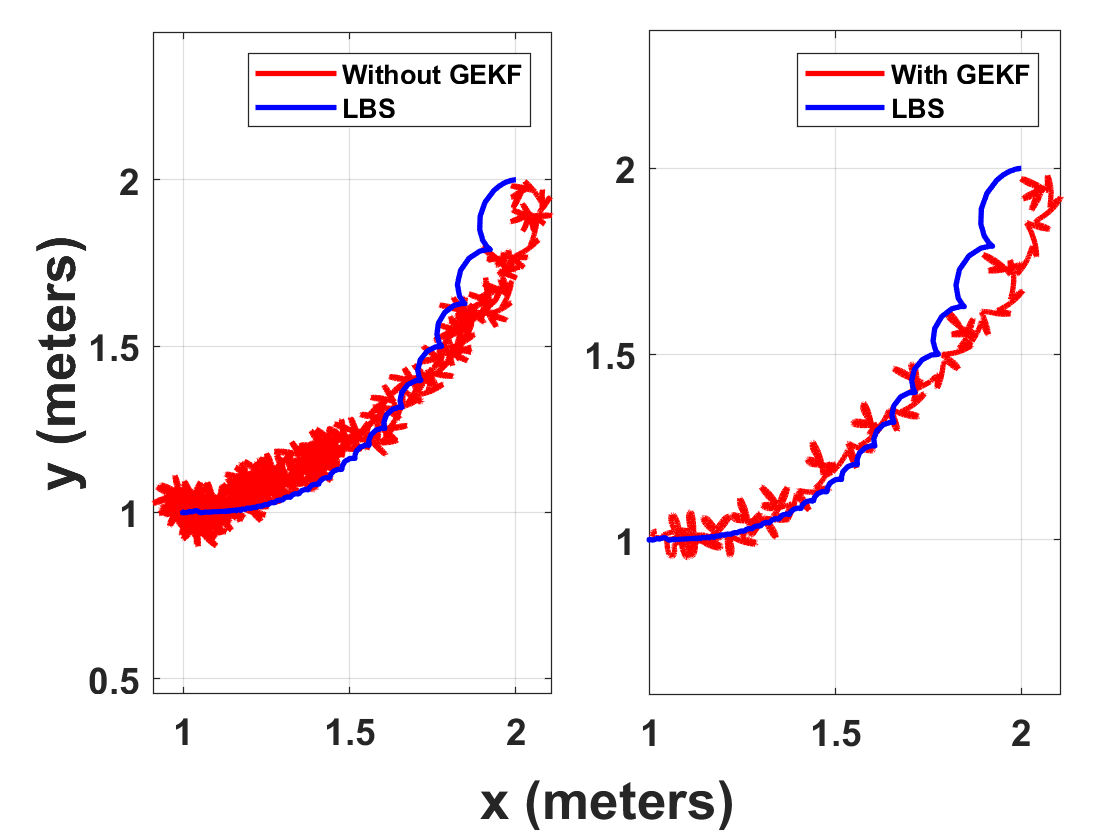}
        
    \end{minipage}
    \caption{Experimental results with known objective. Proposed design (red) converges faster with fewer oscillations than literature baseline design (\cite{durr2013lie}) (blue).}
    \label{fig:ex_R}
\end{figure*}
\begin{figure*}[t]
    \centering
    \begin{minipage}[t]{0.3\textwidth}
        \centering
        \includegraphics[width=\textwidth]{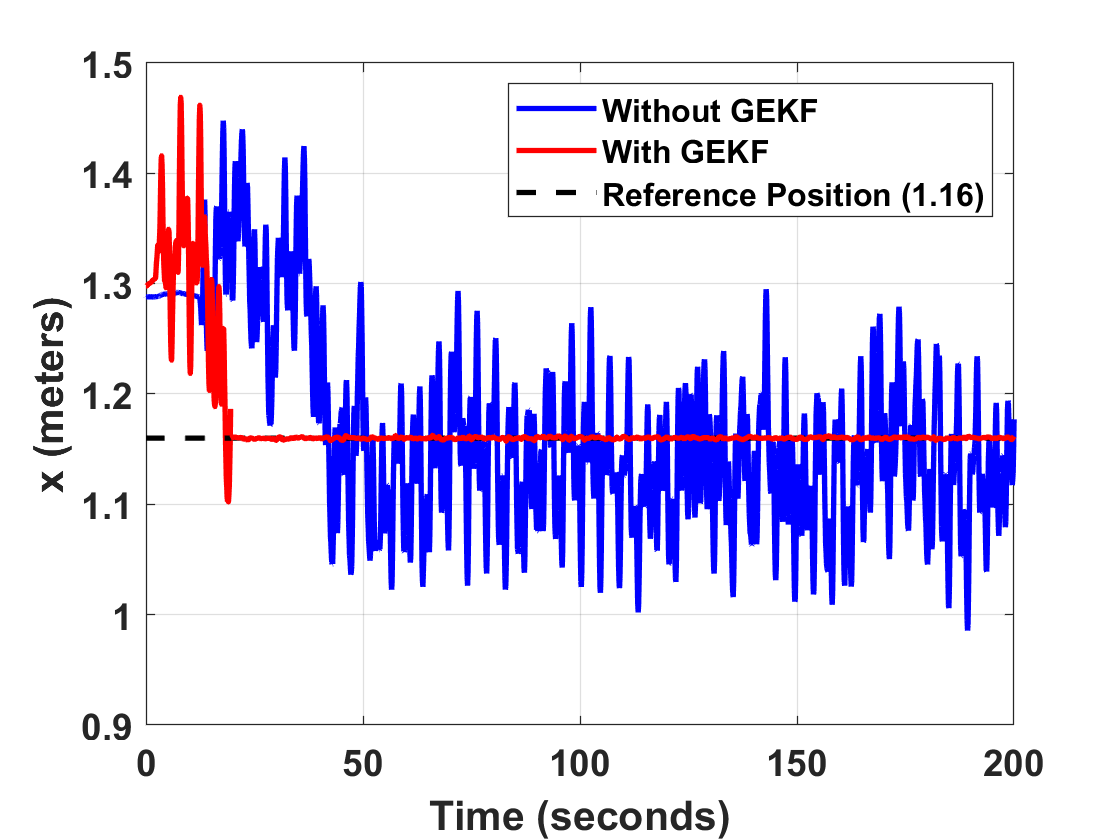}
        
    \end{minipage}
    \begin{minipage}[t]{0.3\textwidth}
        \centering
        \includegraphics[width=\textwidth]{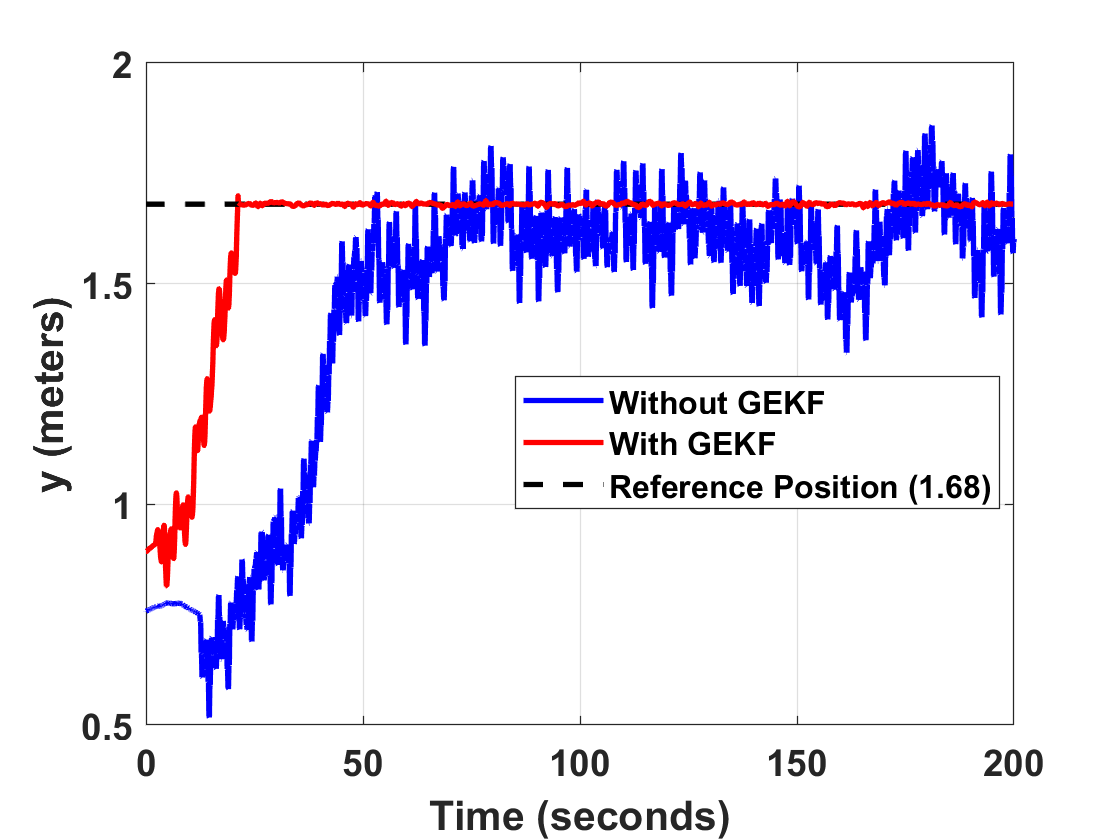}
        
    \end{minipage}
    \begin{minipage}[t]{0.3\textwidth}
        \centering
        \includegraphics[width=\textwidth]{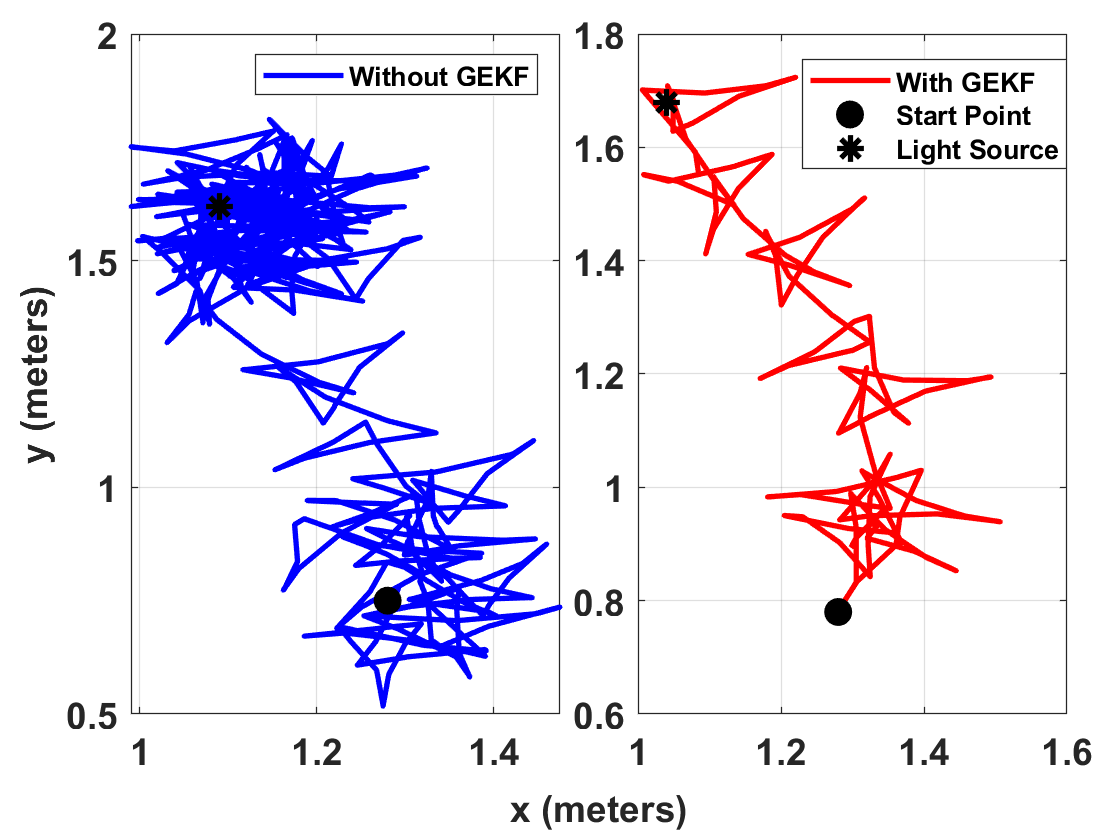}
        
    \end{minipage}
    \caption{Real-time light source seeking. Proposed design (red) converges faster with smoother trajectory than literature baseline design (\cite{durr2013lie}) (blue).}
    \label{fig:ex_R_light}
\end{figure*}
It is important to note that our unicycle-based ESC design (Figure~\ref{fig:ESC_scheme}) requires no state integrators, making it well-suited for differential wheeled robots like the TB3, which takes velocity inputs directly. 

\textbf{Some comments.} In our implementation, $\Omega$ is fixed and only $v$ is updated using the simple control-affine ESC law \eqref{eq:simple_law}, which processes sampled sensor measurements of $f(x,y)$ through high-pass filtering and sinusoidal excitation, enabling a simple, model-free, real-time control framework. Although this ESC law \eqref{eq:simple_law} was previously considered not as effective choice due to persistent oscillations and poor experimental performance~(\cite{grushkovskaya2018class,grushkovskaya2018family}), our recent works~(\cite{pokhrel2023control,ECC_2024}) showed that pairing this law with GEKF (\cite{pokhrel2023gradient}) and an adaptation law to attenuate the amplitude of the excitation signals makes it an effective choice. 

Although there are more complex ESC designs (\cite{grushkovskaya2018class,grushkovskaya2018family,yilmaz2024unbiased}) that can achieve attenuation of oscillation, they often require intricate data processing due to layers of compositions of maps involving the measurements of the objective function; this can be potentially challenging, especially when applied to noisy sampled measurements. Thus, we think it is a promising result to demonstrate that our approach works effectively even when using the very simple ESC law \eqref{eq:simple_law}. 
\section{Simulation and Experimental Results}
In this section, we provide the simulation and experimental results for our proposed unicycle-based ESC design, as presented in Section~\ref{Sec:Main_results}. All simulations are performed using MATLAB\textsuperscript{\textregistered}. For the experimental validation presented in this section, we used the TB3 robot in MDCL lab as shown in \cite{ECC_2024}. We start with the simulation results using a known mathematical expression of the objective function \( f(x,y) = 10 - \frac{1}{2}(x - 1)^2 - \frac{3}{2}(y - 1)^2 \), which attains its maximum at \( (x, y) = (1, 1) \). Simulations were performed in MATLAB and Simulink\textsuperscript{\textregistered} with initial conditions \( (x, y) = (2, 2) \), and GEKF parameters \( P = [4,4,4,4,4]^T \), \( \Delta t = 0.01 \), \( \bm{Q} = 0.0005\bm{I} \), \( R = 0.00999 \), \( N = 10 \). ESC parameters were \( \omega = 120 \), \( c = 0.3 \), \( \Omega = 3 \text{ rad/s} \), \( a(0) = 1 \), \( \lambda = 0.0105 \), and HPF constant \( h = 1 \). Figure~\ref{fig:Simulation_R} shows that with GEKF, the design converges faster and with less oscillation compared to the baseline ESC in \cite{durr2013lie} (no GEKF), clearly validating the theoretical performance advantages.

Moving forward, we extended our study to real-world, real-time experiments using the same known objective function as in the simulations to obtain measurements. We implemented Algorithm~\ref{alg:alg1} on the TB3 robot. The used parameters were \( (x, y) = (2, 2) \), \( P = [4,4,4,4,4]^T \), \( \Delta t = 0.1 \), \( \bm{Q} = 0.0005\bm{I} \), \( R = 0.5 \), \( N = 20 \), \( \omega = 50 \), \( c = 5 \), \( \Omega = 1.75 \text{ rad/s} \), \( a(0) = 3.53 \), \( \lambda = 0.01 \), \( h = 1 \). As shown in Figure~\ref{fig:ex_R}, the robot reached \( (1,1) \) in about 350 seconds using our design, compared to ~2000 seconds with the literature baseline. Also the planner trajectory of the proposed design is captured better by the theoretical LBS when compared to the literature baseline trajectory. This validated faster convergence and oscillation attenuation. Slight variations after convergence are attributed to noise in the motion capture system. The reader can see the experiment video in YouTube (\cite{Unicycle_known_mathamtical_expression_with_GEKF}). 

We further tested the design in a real-time, model-free light source-seeking task where the light intensity
distribution was used as the unknown objective function, with measurements obtained
via a light sensor. The robot used the light sensor to detect intensity, with no model of the sensor, robot, or field. Parameters were \( (x, y) = (1.3, 0.85) \), \( P = [4,4,4,4,4]^T \), \( \Delta t = 0.1 \), \( \bm{Q} = 0.005\bm{I} \), \( R = 0.2 \), \( N = 20 \), \( \omega = 100 \), \( c = 1 \), \( \Omega = 1 \text{ rad/s} \), \( a(0) = 0.55 \), \( \lambda = 0.02 \), \( h = 6 \). The proposed design reached the source in under 24 seconds and stopped. In contrast, the literature design exhibited noisy, prolonged oscillations (Figure~\ref{fig:ex_R_light}). The
corresponding YouTube video showcasing this result can be accessed at (\cite{Unicycle_light_source_with_GEKF}).

\begin{figure}[ht]
    \centering
    \includegraphics[width=0.8\linewidth]{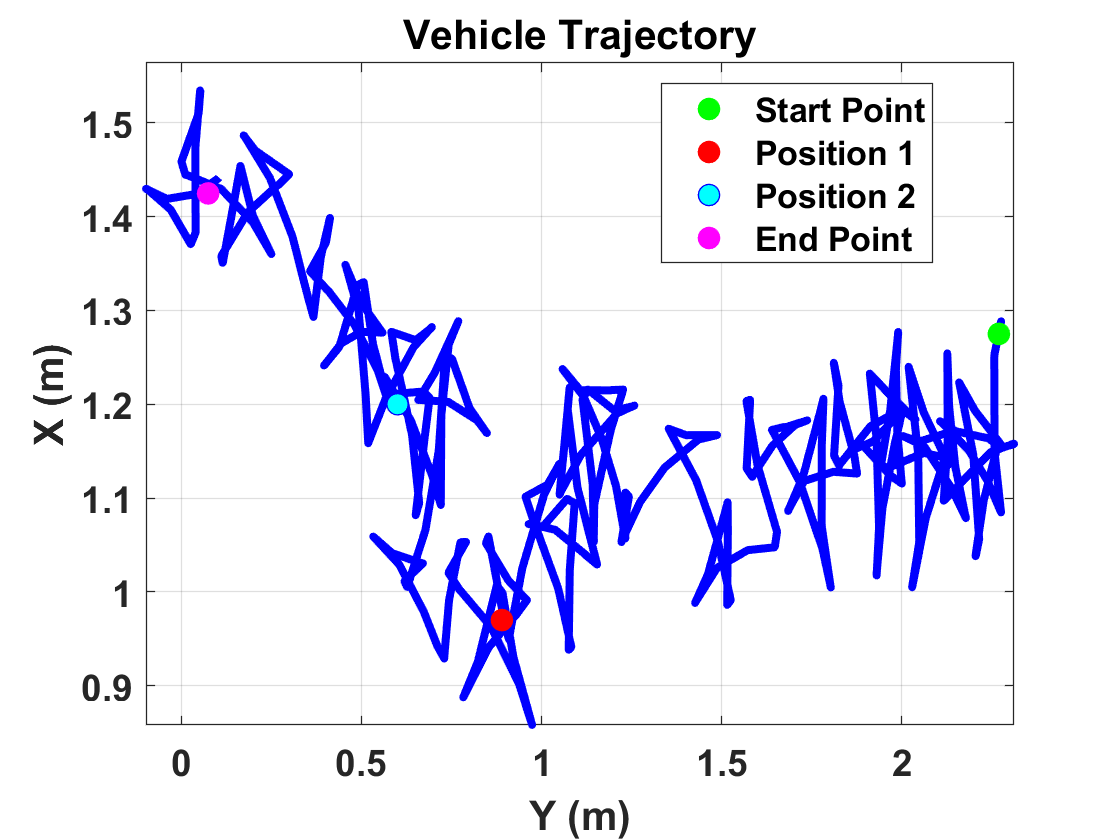}
    \caption{Planar trajectories showing the robot tracking a moving light source over three position changes using the proposed design.}
    \label{fig:ex_L_multiple}
\end{figure}

To test adaptability, we moved the light source three times during a source-seeking task. The proposed design adjusted its excitation via the adaptation law to re-initiate tracking when needed. As shown in Figure~\ref{fig:ex_L_multiple}, the robot successfully followed the source as it changed position as shown in the video at \cite{Unicycle_multiple_positions}.

\textbf{Comparison with the results in \cite{ECC_2024}.} We compare our unicycle-based ESC desvign with the single-integrator-based ESC from~\cite{ECC_2024}. For the known objective function, the ESC design in \cite{ECC_2024} took ~40,000 seconds to converge, while the proposed design took only 350 seconds. In the model-free light source task, the ESC design in \cite{ECC_2024} needed ~3000 seconds, while the proposed design needed 24 seconds to reach the source. Additionally, our approach is more compact, using one ESC law and one adaptation law (Eq.~\eqref{eqn:extended_system}) versus two separate ESC and adaptation laws in the single-integrator-based design in \cite{ECC_2024}. This demonstrates superior performance and simplicity of the proposed design.

\section{Conclusions and Future Work}
In this paper, we proposed a simple and effective unicycle-based ESC design for model-free, real-time source-seeking design, applicable to autonomous robotic systems, particularly differential wheeled robots. The proposed ESC approach, attenuates oscillations, improves convergence speed, and offers better noise handling. These advantages were validated through simulations and real-world experiments. Our design outperformed both classical unicycle ESC methods and our recent single-integrator-based ESC design~\cite{ECC_2024}. Future directions include extending this framework to 3D, adding obstacle avoidance, and applying it to flying robots (e.g., UAVs).

\bibliography{ESC_ref}
\end{document}